\documentclass[letterpaper, 10 pt, conference]{ieeeconf}  

\IEEEoverridecommandlockouts                              

\usepackage{amssymb}  
\usepackage{multirow}

\usepackage{marvosym}
\usepackage{makecell}
\usepackage{color}

\usepackage{graphics} 
\usepackage{epsfig} 
\usepackage{times} 
\usepackage{stfloats}
\usepackage{enumerate}
\usepackage{subcaption}
\usepackage{url}
\usepackage{hyperref}

\newcommand{\caut}[0]{{{\sc cautious}}}
\newcommand{\risk}[0]{{{\sc risky}}}
\newcommand{\nota}[0]{{{\sc not-aligned}}}
\newcommand{\demo}[0]{{{\sc demonstrations}}}
\newcommand{\feed}[0]{{{\sc feedback}}}
\newcommand{\mult}[0]{{{\sc multi-alignment}}}

\title{\LARGE \bf
Driving Style Alignment for LLM-powered Driver Agent
}

\author{Ruoxuan Yang, Xinyue Zhang, Anais Fernandez-Laaksonen, Xin Ding and Jiangtao Gong\textsuperscript{\Letter}
\thanks{The authors are with the Institute for AI Industry Research, Tsinghua University, Beijing, China. Corresponding Email:
        {\tt\small gongjiangtao@air.tsinghua.edu.cn}}
}

\begin{document}

\maketitle
\thispagestyle{empty}
\pagestyle{empty}

\begin{abstract}


Recently, LLM-powered driver agents have demonstrated considerable potential in the field of autonomous driving, showcasing human-like reasoning and decision-making abilities.
However, current research on aligning driver agent behaviors with human driving styles remains limited, partly due to the scarcity of high-quality natural language data from human driving behaviors.
To address this research gap, we propose a multi-alignment framework designed to align driver agents with human driving styles through demonstrations and feedback.
Notably, we construct a natural language dataset of human driver behaviors through naturalistic driving experiments and post-driving interviews, offering high-quality human demonstrations for LLM alignment.
The framework's effectiveness is validated through simulation experiments in the CARLA urban traffic simulator and further corroborated by human evaluations.
Our research offers valuable insights into designing driving agents with diverse driving styles.
The implementation of \href{https://github.com/AIR-DISCOVER/Multi-alignment-Drivng-Agent}{the framework}\footnote{\url{https://github.com/AIR-DISCOVER/Multi-alignment-Drivng-Agent}} and details of \href{https://github.com/AIR-DISCOVER/Driving-Thinking-Dataset}{the dataset}\footnote{\url{https://github.com/AIR-DISCOVER/Driving-Thinking-Dataset}} can be found at the link.

\end{abstract}

\section{INTRODUCTION}
In the burgeoning field of autonomous driving (AV), driver agents powered by Large Language Models (LLMs) are demonstrating remarkable promise due to their exceptional planning\cite{song2023llm} and reasoning\cite{wei2022chain,wang2022self,yao2023tree} capabilities.
Researchers have delved into the development of intricately designed driver agents that could perceive environmental stimuli\cite{chen2023driving,xu2023drivegpt4,hu2023gaia}, comprehend the situation\cite{shao2023lmdrive}, fetch their memories\cite{cui2024drive,wen2023dilu} and deduce subsequent driving actions\cite{wang2023drivemlm} that mirrors human decision-making.
Such human-like AVs show promise in navigating a diverse range of driving scenarios~\cite{kolekar2020human,jin2023surrealdriver}, enabling better anticipation of AV behavior by other road users~\cite{hecker2019learning}, while also enhancing human trust in these systems~\cite{waytz2014mind}.

However, aligning these driver agents with human driving styles to imbue them with more human-like and personalized characteristics remains unexplored.
Prevailing strategies for aligning LLM-based agents with humans, such as fine-tuning\cite{chen2023driving,xu2023drivegpt4,mao2023gpt} and the integration of expert feedback\cite{ouyang2022training,fu2024drive}, are often hindered by their high costs.
Recently, some studies have leveraged AI to generate feedback or reflections\cite{zhao2023expel,shinn2024reflexion,yao2023retroformer,yang2023failures}, yet they fall short in aligning such reflections with human perspectives.
On the other hand, despite researches focusing on employing AI to generate few-shot demonstrations\cite{song2023llm,wang2023large} for LLMs, another challenge in enhancing agent-human alignment lies in the lack of high-quality human behavior data in a form accessible to LLMs, making it difficult for agents to learn from human demonstrations.
Existing datasets for autonomous driving learning either provide only environment data for perception tasks\cite{caesar2020nuscenes,geyer2020a2d2,huang2018apolloscape} rather than driving behaviors, or present driving behaviors in non-linguistic modalities (e.g. trajectories in maps\cite{zhan2019interaction}, Controller Area Network Bus (CAN-Bus) data\cite{li2019driving,hu2022processing}, in-car videos\cite{martin2019drive}) that are indirect for LLMs to learn from.
Thus, successful alignment requires an approach that efficiently synchronizes LLM-based driver agents with human driving styles, as well as a collection of driving demonstrations across different driving styles in natural language for LLMs' comprehension and learning.

In this paper, we introduce a novel multi-alignment framework that utilizes demonstrations and feedback to align driver agents with human driving styles.
Diverging from reliance on human expert feedback or reflections from LLMs themselves, our approach harnesses the few-shot learning capabilities\cite{brown2020language} of LLMs to create a Coach Agent that learns from human demonstrations, evaluates past driving behaviors, and formulates driving guidelines.
All human demonstrations are pre-collected, eliminating the need for additional human effort during alignment and substantially reducing costs.

Moreover, to collect high-quality demonstrations for alignment, we compiled a dataset that encompasses driving behaviors from drivers with varied driving styles.
A real-world driving experiment was conducted, followed by a post-driving interview, wherein we gathered and structured human drivers' decision-making data.
This dataset likely represents the first effort to meticulously dissect human driving behaviors and articulate the driving decision-making process in a natural language format.

We validate our work through both simulation experiments and human evaluation surveys, demonstrating that our multi-aligned framework effectively creates driver agents with distinct driving styles that are not only statistically sound but also distinctly perceptible to humans.

The contributions of this paper are summarized as follows:

\begin{itemize}
    \item A multi-alignment framework that can align LLM-based driver agents with human driving styles.
    \item A dataset of human driving behaviors in natural language format.
    \item Comprehensive validation through both simulation experiments and human evaluations.
\end{itemize}

\section{Multi-alignment Framework}

\begin{figure*}
\centering
\includegraphics[width=\textwidth]{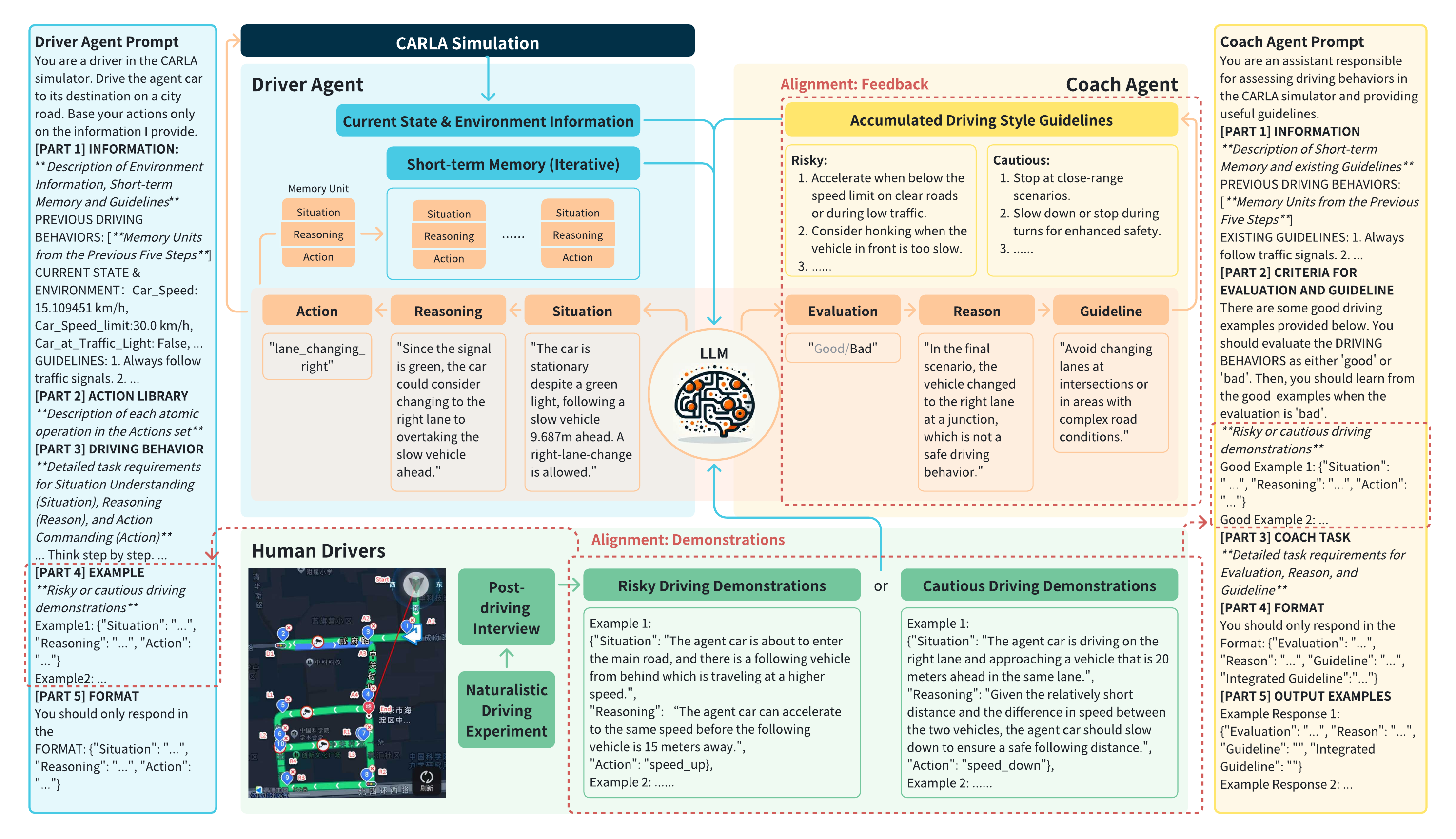} 
\caption{The multi-alignment framework} 
\label{fig:mainimage}
\end{figure*}

Fig. \ref{fig:mainimage} demonstrates the comprehensive structure of the multi-alignment framework, consisting of a Driver Agent, a Coach Agent, and demonstrations from human drivers. 
In this section, we first introduce the architecture and basic workflow of the Driver Agent.
Then we show how to achieve multi-alignment through direct demonstration data from human drivers and feedback from the Coach Agent with human demonstrations.

\subsection{Driver Agent}
The Driver Agent acts as entities interacting with the surrounding driving environment and making driving decisions.
It maintains an iterable, fixed-capacity short-term memory, which stores the most recent memory units, promoting the continuity and consistency of decision-making.

The workflow begins by capturing the current state and environment information for perception, including the speed and direction of the agent vehicle, the speed limits and other restrictions of the current road, as well as the status of other vehicles and pedestrians nearby.
Next, it analyzes the collected information alongside its short-term memory to grasp the current situation.
Following this analysis, along with provided Demonstrations and Guidelines for multi-alignment, the Driver Agent deduces the most appropriate driving action at the moment.
Here, the Driver Agent is prompted to 'Think Step by Step,' employing a chain-of-thought (CoT) reasoning strategy towards the final decision:

``\textit{Given the rather faster speed of the vehicle ahead and inability to change lanes, the agent car should match the speed by gentle acceleration.}''

Next, the Driver Agent selects the most matching ones from a set of atomic driving operations as the step's action and performs.
The "Situation," "Reasoning," and "Action" generated are then compiled into a memory unit and incorporated into the short-term memory, while the earliest memory unit is popped out.
Through the consistent repetition of this process, the Driver Agent successfully crafts a sequence of fluid and coherent driving maneuvers.

\subsection{Multi-alignment through Demonstrations and Feedback}
\label{sec:multi-alignment}

We construct a framework that could multi-align the Driver Agent with human driving styles by adopting demonstrations and feedback.

Demonstrations encompass representative decision-making processes of human drivers, featuring both cautious and risky driving demonstrations.
They are collected and then organized into the form of the Driver Agent's memory units (with more details in Section \ref{sec:data collection}).
Demonstrations serve a dual purpose in alignment, being utilized by both the Driver Agent and the Coach Agent.
For the Driver Agent, they serve as few-shot prompts, aiming to guide it towards making driving decisions similar in style.
And for the Coach Agent, they are provided as 'Good' examples, prompting it to make evaluations with driving style preferences, further generating guidelines that embody driving styles.

To implement feedback, a Coach Agent was established, outfitted with a Guidelines module that compiles driving suggestions gleaned from continuous evaluations.
It scrutinizes the current short-term memory of the Driver Agent and issues a judgment of 'Good' or 'Bad', along with the reason for this judgement.
The criteria for evaluation include whether the decisions in the short-memory align with common driving sense, conform to the requirements proposed in the Guidelines, and match the style of the provided 'Good' examples.
Should an evaluation yield a 'Bad' rating, the Coach Agent formulates a new guideline addressing the suboptimal driving decision.
This new guideline is then assimilated into the existing Guidelines repository.

\section{Driving Style Data Collection}
\label{sec:data collection}

\subsection{Natural Driving Experiment and Post-driving Interview}

To gather authentic human driving behavior data for alignment, we conducted a natural driving experiment with human drivers followed by a post-experiment interview.
A total of 24 drivers were invited to participate in our data collection experiment, covering different genders and age groups. Notably, in order to gather data on different driving styles, the participants also included both seasoned professional drivers and novice drivers with less driving experience.

To delve deeply into specific driving behaviors, we initially had each driver perform an urban road driving task covering 13 driving conditions, with a total length of 5.7 kilometers.
To faithfully recreate the entire driving process during the following post-experiment interview, we set up a roof-mounted 360-degree panoramic camera to record the environment around the vehicle during task execution, an in-car motion camera to capture the driver's actions, as well as an eye tracker to record changes in the driver's gaze.
Additionally, real-time CAN-Bus data on the vehicle's status were recorded, including speed, the throttle and brake percentage, and the turning of the steering wheel.

For safety reasons, drivers were not requested to verbalize their thought processes while performing driving tasks.
Right after the natural driving experiment, drivers would participate in a detailed post-experiment interview, which typically lasted for 1.5-2 hours.
During the interview, we used the collected videos to recreate the task situation just experienced by the driver. 
For each driving action (e.g. accelerating, lane changing or turning), drivers were asked to recall and describe the entire decision-making process, from evaluating the surrounding environment to executing the corresponding driving action.
These interview data will assist in determining the driver's driving style, and also serve as the source of Demonstrations in the Multi-alignment Framework.

\subsection{Driving Style Selection and Data Organization}

Having completed driving experiments and post-experiment interviews, our next task is to differentiate the drivers' driving styles and organize the think-aloud data into demonstrations of different styles.

The differentiation of driving styles is based on subjective questionnaire results and objective driving records in driving tasks.
We distributed a MDSI questionnaire\cite{taubman2004multidimensional} to each driver invited to participate in the experiment.
The results indicated the presence of four driving styles among the 24 drivers: risky, high-velocity, patient, and careful.
Notably, the risky style often coincided with the high-velocity style, while the patient style typically appeared alongside the careful style.
Further analysis of the CAN-Bus data during driving tasks revealed that 3 drivers exhibited speeds and throttle percentages significantly above the average — specifically, the average speed of all drivers was 6.40 m/s and average throttle percentage was 23.09\%, while average speed of these 3 drivers respectively reached speeds of 7.73 m/s (20.78\% higher than average), 7.50 m/s (17.19\% higher than average) and 7.41 m/s (15.78\% higher than average), and average throttle percentages reached 29.09\% (25.99\% higher than average), 24.42\% (5.76\% higher than average) and 24.37\% (5.54\% higher than average) — aligning with their self-reported 'risky and high-velocity' driving styles in the  questionnaire.
Conversely, 2 other drivers had lower metrics — with speeds of 5.15 m/s (19.53\% lower than average) and 5.28 m/s (17.50\% lower than average) respectively, and throttle percentage of 21.00\% (9.05\% lower than average) and 21.34\% (7.58\% lower than average) — aligning with their self-reported 'patient and careful' driving styles in the  questionnaire.
Additionally, a few drivers who reported to have driving styles in the questionnaire did not show clear trends in either driving data or interview records.

Therefore, we identified two basic driving styles: 'risky' and 'high-velocity' were merged into 'risky,' while 'patient' and 'careful' were combined into 'cautious.'
We reviewed the interview data of drivers with risky driving styles and those with cautious driving styles, selecting representative decision-making processes that exemplify each driving style.
Then, we organized each process according to the decision sequence into the format of 'Situation', 'Reasoning' and 'Action', forming the final Demonstrations for alignment with humans.

\section{Experiment}

In this section, we validated the proposed Multi-alignment Framework by exploring the following questions:

\begin{itemize}
    \item Can Driver Agents with different driving styles be constructed using human think-aloud data?
    \item Which alignment method can efficiently achieve human alignment of driving styles?
\end{itemize}

To this end, we implemented the Multi-alignment Framework on CARLA—a high-fidelity traffic simulator. A simulation experiment was carried out under urban driving conditions, upon which we further conducted a user experiment to collect human evaluations of its performance.

\subsection{Conditions}

\begin{figure}
\centering
\includegraphics[width=1.0\linewidth]{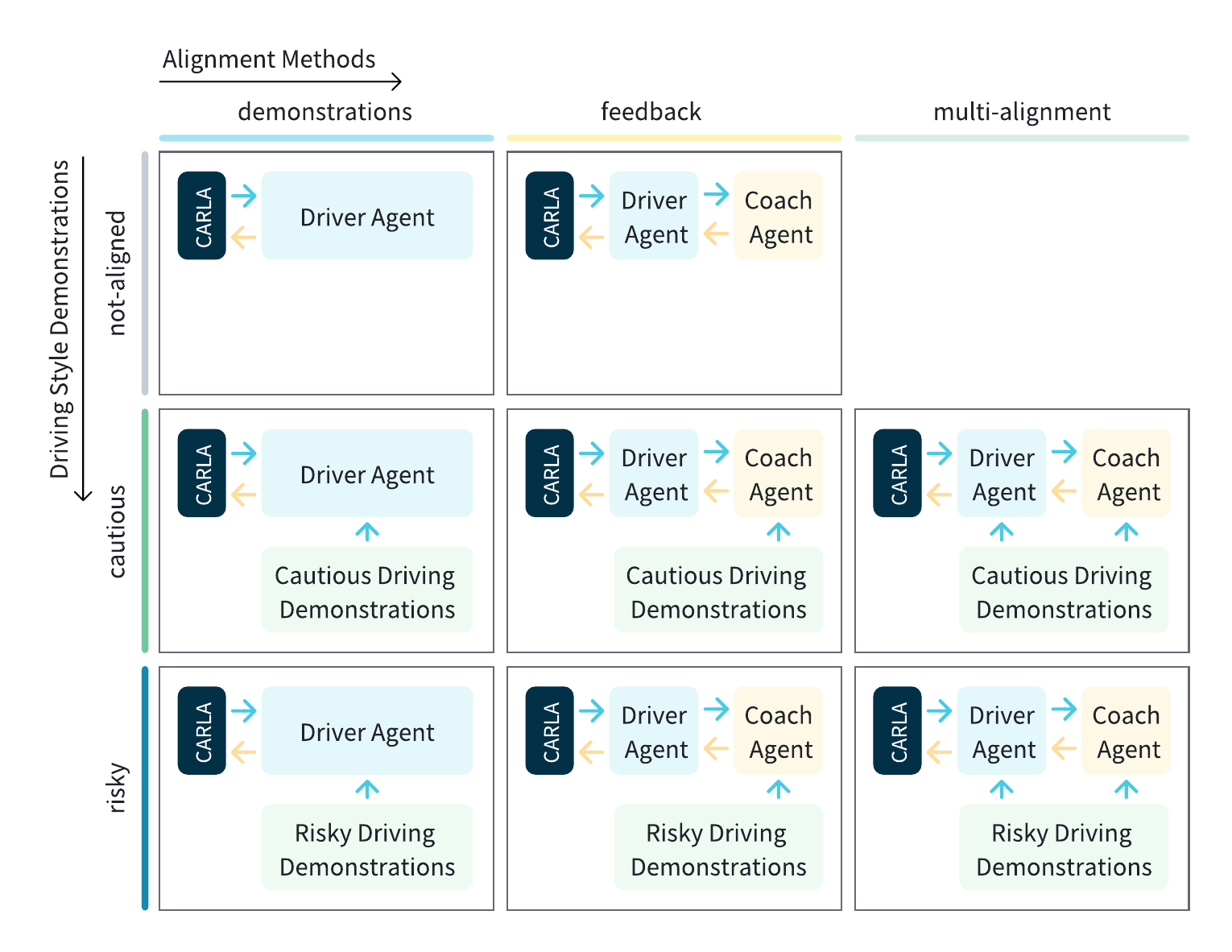} 
\caption{Experiment conditions} 
\label{fig:conditionimage}
\end{figure}

The experiment adopts an approximate 3 × 3 with-in subject design with two main variables:
Driving Style [\caut\ (C), \risk\ (R)and \nota\ (N)] and Alignment Method [\demo\ (D), \feed\ (F)and \mult\ (M)]. Fig. \ref{fig:conditionimage} shows the general design of different conditions.

In terms of Driving Style, we compared the effects of using \caut\ driving demonstrations, \risk\ driving demonstrations, and no demonstrations (i.e., \nota).

Alignment Method was organized in an ablation format, with conditions including \demo, \feed, and \mult\ (i.e., the full alignment framework).
The \demo\ condition involves Driver Agents provided with demonstrations, and the \feed\ condition involves Driver Agents without demonstrations and Coach Agents that were provided with demonstrations, while in the \mult\ condition, both Driver Agent and Coach Agent were provided with demonstrations.

\subsection{CARLA Simulation}

\subsubsection{Set-up}

The simulation experiment setup involved a ThundeRobot Zero desktop computer as the hardware foundation. The simulation environment was built upon the CARLA simulator, specifically, version 0.9.14\footnote{\url{http://carla.org/2022/12/23/release-0.9.14/}} and operated on Python 3.7 with Unreal Engine 4\footnote{\url{https://docs.unrealengine.com/4.27/en-US/}}.
We use the map Town10, a typical urban driving scene, with both the number of other vehicles and pedestrians in the scenario set to 60. And Audi TT was the designated vehicle for all experiments, with fixed starting and continuously, randomly generated ending points for its path (After a vehicle is generated at a predefined fixed point, a random endpoint is generated. Upon reaching the endpoint, another endpoint is randomly generated, and so on.).

We leverage OpenAI’s GPT-4\footnote{\url{https://openai.com/gpt-4}} APIs for constructing both the Driver Agent and the Coach Agent. However, it takes several seconds for GPT to generate a response, which is too long in a driving context for making immediate decisions. Therefore, we slowed down CARLA’s simulation time based on the required token count by setting a fixed time-step of 0.0008-0.0015 seconds.

Each simulation process is recorded on video.
Additionally, to collect vehicle status information during the simulation, we initiated a log-collector thread to accumulate log on collisions, speed, throttle percentage, and brake percentage from the agent vehicle on a second-by-second basis.

\subsubsection{Metrics}

Here, we introduce three metrics to evaluate the driving performance of the Driver Agent: collision rate, speed, throttle percentage, and brake percentage.

\begin{itemize}
    \item \textit{Collision rate:} The number of collisions can be obtained from the log, with distance traveled being cumulative up to the last collision.
    The calculating formula is
    $Collisions\ per\ Meter=\frac{Number\ of\ Collisions}{Distance\ Traveled\ (m)}$
    \item \textit{Speed:} The statistical measures for speed include the average speed of the agent vehicle during each simulation and the segmented average speed per minute (simulator time).
    All calculations of average speed exclude zero values to minimize the impact of the agent vehicle waiting at traffic signals and in traffic jams.
    \item \textit{Throttle percentage \& brake percentage:} The statistics for throttle and brake percentages are also divided into overall average values and segmented average values per minute. Similarly, all calculations exclude data from when the agent vehicle is stationary.
\end{itemize}

\subsubsection{Results}

\begin{figure*}[t]
    \centering
    \begin{subfigure}{.32\textwidth}
        \centering
        \includegraphics[width=\linewidth]{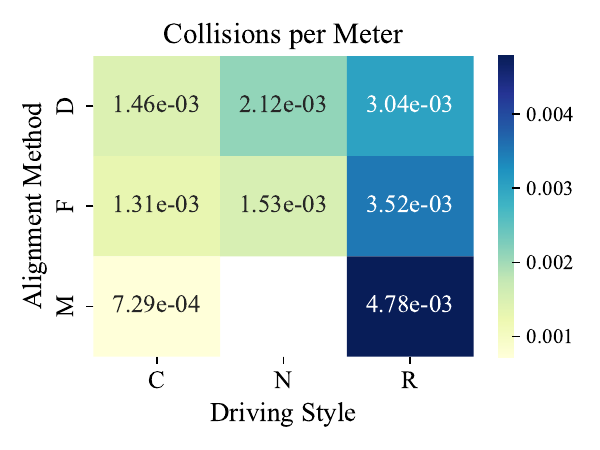}
        \subcaption{Collision rates per meter (with increased incidences of abrupt maneuvers by surrounding vehicles and pedestrians).}
        \label{chart:sub1}
    \end{subfigure}%
    \hspace{10pt} 
    \begin{subfigure}{.63\textwidth}
        \centering
        \includegraphics[width=\linewidth]{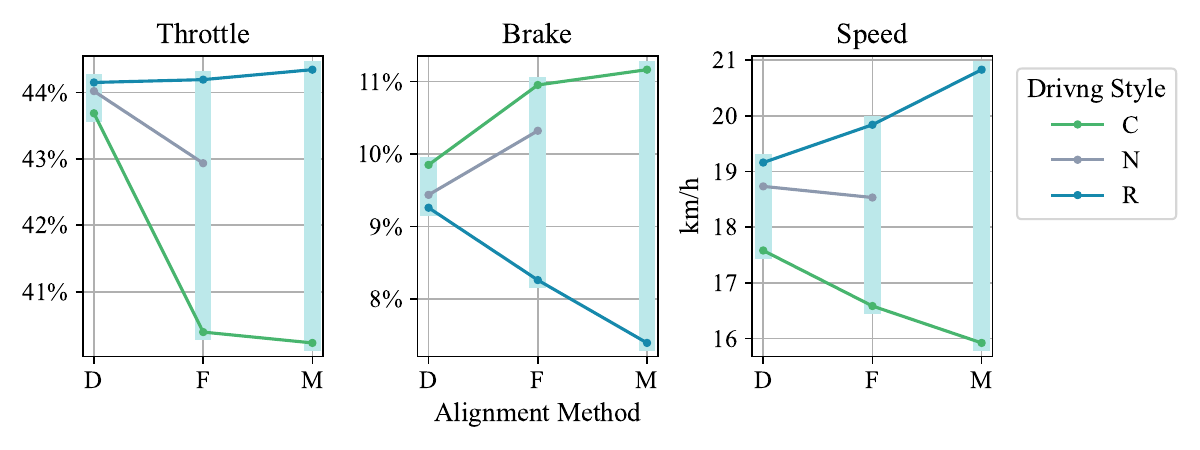}
        \subcaption{Average throttle percentage (left), brake percentage (middle), and speed (right) of the agent vehicle, with all calculations excluding data from when the agent vehicle was stationary (the speed limit is km/h).}
        \label{chart:sub2}
    \end{subfigure}
    \caption{Simulation experiment results for predefined metrics. }
    \label{fig:test}
\end{figure*}

We conducted approximately 50.3 hours of simulation experiments under various conditions, which corresponds to an average of about 6.7 minutes of driving per condition for the agent vehicle on the simulation platform.
The average distance the agent vehicle traveled per condition was approximately 1.5 kilometers.
Notably, we adjusted the algorithms controlling other vehicles and pedestrians to make them more prone to sudden maneuvers (e.g. abrupt lane changes, running red lights).
These edge cases aim to increase the risk level of the driving environment for the agent vehicle, making its driving style more observable.

Fig. \ref{chart:sub1} displays the collision rates per meter for the agent vehicle calculated under different conditions.
Agents aligned with the \risk\ driving style overall exhibit higher collision rates, while those aligned with the \caut\ driving style show lower collision rates overall.
Additionally, when aligned with \caut\ driving style, the \mult\ method displayed the lowest collision rate while the \demo\ method displayed the highest, and when aligned with \risk\ driving style, the \mult\ method showed the highest collision rate while the \demo\ method displayed the highest.
When \nota, the collision rate for the \demo\ method is higher than that for the \feed\ method.

Fig. \ref{chart:sub2} presents the average throttle percentage, brake percentage, and speed of the agent vehicle during the driving process under different conditions, with all calculations excluding data from when the agent vehicle was stationary.
When using the same alignment method, agents aligned with the \risk\ driving style had the highest average speed, highest throttle percentage, and lowest brake percentage, while agents aligned with the \caut\ driving style had the lowest speed, lowest throttle percentage, and highest brake percentage.
When aligned with the \caut\ driving style, the average speed and throttle percentage decrease while the average brake percentage increases across the \demo, \feed, and \mult, in that order. The opposite trend is observed when aligning with the \risk\ driving style.
When \nota, the average speed and throttle percentage for the \demo\ method are higher than those for the \feed\ method, while the average brake percentage is lower.

\subsubsection{Findings} Based on the hypothesis that agents with more cautious driving styles are safer, agents can exhibit corresponding driving styles by aligning with different driving styles.
\mult\ was the most effective method, displaying the most significant differences in collision rates, average throttle, brake, and speed between cautious and risky driving styles, while \demo\ were less effective.

\subsection{Human Evaluation}

\subsubsection{Procedure}

We designed two survey questionnaires to collect human drivers' evaluations of the Driver Agent's performance, which was presented to participants in the questionnaire through video clips of the simulation, with about 30 seconds of driving footage captured for each experimental condition.


In Part I of the first questionnaire, we initially collected basic information (e.g. age, gender, whether holding a driving license) from participants. A partial MDSI self-assessment was also included, with items covering indicators of risky and careful driving styles from the MDSI.

In Part II, the video clips are divided into four groups:
\begin{itemize}
    \item Demonstrations Group: \{\demo\ \caut\ (DC), \demo\ \nota\ (DN), \demo\ \risk\ (DR)\}
    \item Feedback Group: \{\feed\ \caut\ (FC), \feed\ \nota\ (FN), \feed\ \risk\ (FR), \demo\ \nota\ (DN, serving as baseline in this group)\}
    \item Cautious Group: \{\demo\ \caut\ (DC), \feed\ \caut\ (FC), \mult\ \caut\ (MC)\}
    \item Risky Group: \{\demo\ \risk\ (DR), \feed\ \risk\ (FR), \mult\ \risk\ (MR)\}
\end{itemize}
Each group of video clips will appear in a random order, accompanied by a ranking question requiring participants to rank the driving styles in the videos according to their level of riskiness (a smaller number indicates more risky) and a reason question for explaining their rankings.

Parts I of the second questionnaire are identical to the first questionnaire.
In Part II, participants were instructed to watch all of the eight videos clips, which were also organized in a random order, with three scoring questions respectively investigated the intelligence level, riskiness level and human-likeness level of the agent vehicle (all from 0 to 10) and a reason question attached below each clip.


Additionally, to filter out carelessly completed questionnaires, we set a minimum answering time and included trap questions in the questionnaire, which required participants to select a certain option.

\subsubsection{Participants}

We recruited over 200 participants through a third-party recruitment channel provided by the survey platform, offering a compensation of approximately \$2.08 per valid questionnaire completed.
Additionally, our team of five researchers also shared our questionnaires on social media platforms, recruiting over 60 participants.

All 270 participants verified in the questionnaire that they possess a driving license.
Among them, there were 141 male participants, accounting for 52.22\%, and 129 female participants, accounting for 47.78\%, with ages ranging from 19 to 54 years old.

\subsubsection{Data Analysis}

For both questionnaires, we first categorized participants' driving styles based on the results from Section I. 
The formula for calculating the driving style score is
$S_{driving\ style} = \Sigma o_{risky} -\Sigma o_{cautious}$,
where $o_{risky}$ represents the option made by participants for each risky indicator, while $o_{cautious}$ represents the option for each cautious indicator (with two negative indicators within, where options are included as negative values).
The higher the driving style score, the more a participant's driving style tends towards being risky.

For Part II of the first questionnaire, we calculated the rankings obtained by different video clips in the ranking question following each group of video clips, as well as the statistical significance between their rankings.

For Part II of the second questionnaire, we separately tallied the results of the three scoring questions after each video clip, representing the agent vehicle's intelligence, riskiness, and human-likeness.

Additionally, we scrutinized all the answers to the reasoning questions in both questionnaires, summarizing supports for judging the driving behaviors of the agent vehicles.
\subsubsection{Results}

\begin{figure*}[t]
    \centering
    \begin{subfigure}{.68\textwidth}
        \centering
        \includegraphics[width=\linewidth]{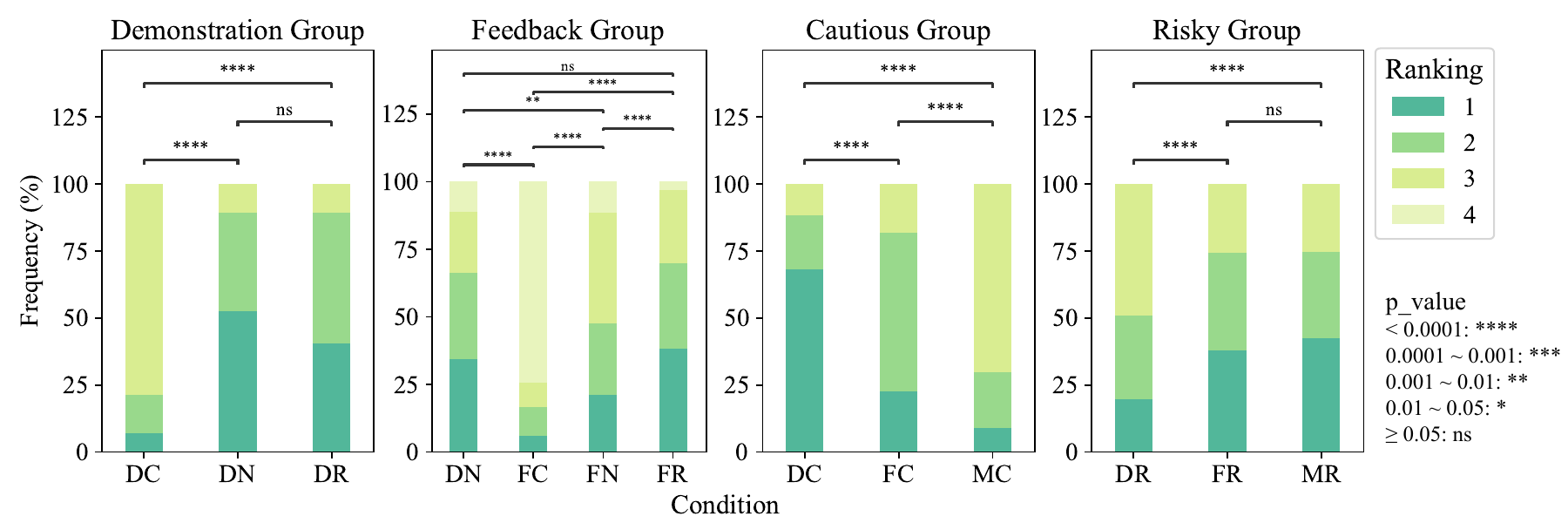}
        \subcaption{Frequency of riskiness rankings in different groups: demonstrations with different driving styles (left), feedback with different driving styles (middle-left), cautious driving style under different alignment methods (middle-right), and risky driving style under different alignment methods (right). Higher rankings indicate higher riskiness.}
        \label{chart:sub3}
    \end{subfigure}%
    \hspace{10pt} 
    \begin{subfigure}{.27\textwidth}
        \centering
        \includegraphics[width=\linewidth]{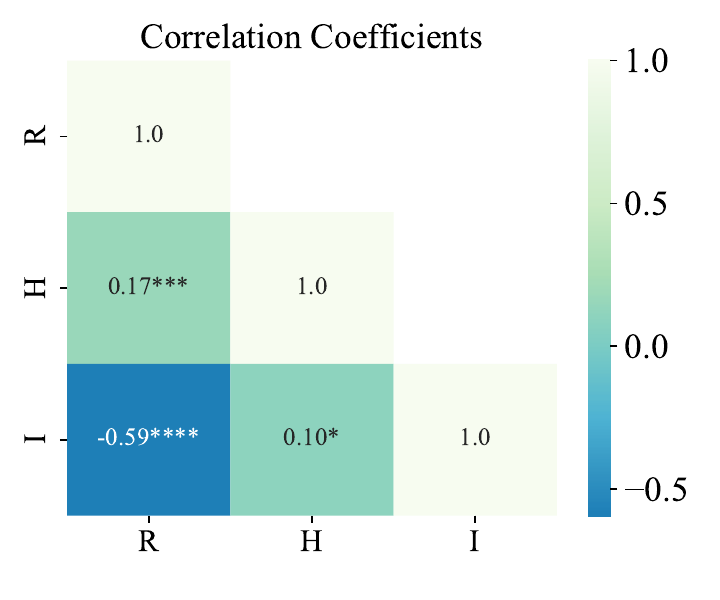}
        \subcaption{Pearson correlation and significance of scores for agent vehicle's riskiness (R), human-likeness (H), and intelligence (I).}
        \label{chart:sub4}
    \end{subfigure}
    \caption{Human evaluation results.}
    \label{fig:test}
\end{figure*}

 We distributed two questionnaires for 3 days and received a total of 259 valid responses after screening, with 198 for the first questionnaire and 59 for the second. 
 The driving style statistics in part I are highly diverse.
 With an average score of 0.61, 34 participants scores below -4 (suggesting a cautious driving style), while 37 participants scores over 5 (suggesting a risky driving style), indicating good representativeness of our results.

 Fig. \ref{chart:sub3} shows the rankings of riskiness for different video clips in each group from the first questionnaire, with higher rankings indicating higher riskiness.
 
 In both the demonstrations and feedback groups, the rankings for DC and FC were significantly lower than those for other videos in the same group, indicating that they were perceived as the least risky.
 One participant explained choosing DC as the least risky in the Demonstration Group, noting, "\textit{The car ran stably without veering left or right.}"
 Another participant cited their reasoning for deeming FC the least risky in the Feedback Group, stating, "\textit{It waits for the pedestrian ahead to pass by.}"
 
 When \nota, the riskiness of FN decreases compared to DN, with multiple participants noting DN's "\textit{Decelerate too slowly when approaching a pedestrian crossing.}" 
 However, DN shows no significant difference when compared to either DR or FR, because they "\textit{all look very risky}"
 
 In the cautious group, the ranking of riskiness goes significantly as DC $>$ FC $>$ MC, indicating that \mult\ has the best alignment effect, with \demo\ being the least effective.
 The majority of participants attributed the rankings to "\textit{Driver x (DC) performs lane changes a bit too quickly, whereas driver y (MC) not only waits for pedestrians but also yields to other vehicles.}"
 
 Similarly, the \demo\ method also showed the poorest alignment effect in the risky group, with MR slightly better than FR but not significant.

Fig.~\ref{chart:sub4} presents the results of the correlation analysis among participants' scores for riskiness, human-likeness, and intelligence for the same video clip in the second questionnaire.
 It can be observed that humans tend to associate higher riskiness with lower intelligence, and higher intelligence with greater human-likeness.
 Interestingly, despite cautious driving being safer, humans still tend to associate higher riskiness with greater human-likeness.
 One participant remarked, "\textit{It (MR) is really like an experienced driver who is showing off his driving skills.}"

\subsubsection{Findings} The human evaluation results indicated a clear distinction in perceived riskiness between different driving styles. 
Agents aligned with cautious driving were consistently rated as less risky, particularly under the multi-alignment condition, which was proven to be the most effective for aligning driving styles. 
Demonstrations alone showed the least effectiveness in both cautious and risky conditions. Additionally, there is an interesting psychological insight that despite associating cautious driving with safety, participants tended to equate higher cautiousness with less human-likeness, reflecting a complex perception of human driving behavior.

\section{CONCLUSIONS}

This paper presents a novel multi-alignment framework for aligning LLM-powered Driver Agents with human driving styles.
Through a comprehensive set of experiments and evaluations, we successfully demonstrate that Driver Agents can be tailored to exhibit distinct driving styles—risky and cautious—by leveraging human driving data as chain-of-thought prompts.
The framework's effectiveness is validated through simulation experiments in the CARLA urban traffic simulator and further corroborated by human evaluations.

By illustrating the potential of LLMs in achieving nuanced human-agent alignment, this work opens new avenues for research into autonomous driving technologies that cater to individual preferences.
By encoding the intricacies of human driving behaviors in a format accessible to language models, this work paves the way for more intuitive and effective human-agent alignment across a broad spectrum of applications beyond autonomous driving.
Additionally, the insights into human perceptions of riskiness and human-likeness in driving styles underscore the complexity of aligning autonomous agents with human expectations and behaviors, highlighting the importance of further interdisciplinary research in this area.




\bibliographystyle{IEEEtran}
\bibliography{IEEEfull,root}

\end{document}